\documentclass[10pt,twocolumn,letterpaper,table]{article}

\usepackage{cvpr}
\usepackage{times}
\usepackage{epsfig}
\usepackage{graphicx}
\usepackage{amsmath}
\usepackage{amssymb}
\usepackage{cite}
\usepackage{adjustbox}
\usepackage{{booktabs}}
\usepackage{multirow}
\usepackage{multicol}
\usepackage{multicolrule}
\usepackage[font=normalsize]{caption}
\usepackage{graphicx}
\usepackage{diagbox}

\newcommand{\tablebar}[2]{{\color{#2}\rule{#1cm}{8pt}}}

\usepackage{enumitem}
\setlist{nosep}
\usepackage{float}

\usepackage[pagebackref=true,breaklinks=true,letterpaper=true,colorlinks,bookmarks=false]{hyperref}
\hypersetup{
    colorlinks=true,
    linkcolor=green,
    filecolor=magenta,      
    urlcolor=blue,
    pdfpagemode=FullScreen,
    }

\cvprfinalcopy 

\def\cvprPaperID{****}

\ifcvprfinal\fi
\begin{document}

\title{Revisiting Simple Baselines for In-The-Wild Deepfake Detection}

\author{
Orlando Castaneda$^{1\ast}$ \hspace{24pt} Kevin So-Tang$^{1}$\thanks{Equal contribution} \hspace{24pt} Kshitij Gurung$^{1}$
\\ \\
$^1$Georgia Institute of Technology
}

\maketitle
\footnotetext[1]{All authors contributed as graduate students at Georgia Institute of Technology}

\maketitle

\begin{abstract}
The widespread adoption of synthetic media demands accessible deepfake detectors and realistic benchmarks. While most existing research evaluates deepfake detectors on highly controlled datasets, we focus on the recently released ``in-the-wild'' benchmark, \textbf{Deepfake-Eval-2024}\cite{chandra_deepfake-eval-2024_2025}. Initial reporting~\cite{chandra_deepfake-eval-2024_2025}  on Deepfake-Eval-2024 showed that three finetuned open-source models achieve accuracies between 61\% and 69\%, significantly lagging behind the leading commercial deepfake detector with 82\% accuracy. Our work revisits one of these baseline approaches, originally introduced by Ojha et al.~\cite{ojha_towards_2023}, which adapts standard pretrained vision backbones to produce generalizable deepfake detectors. We demonstrate that with better-tuned hyperparameters, this simple approach actually yields much higher performance --- 81\% accuracy on Deepfake-Eval-2024 --- surpassing the previously reported accuracy of this baseline approach by 18\% and competing with commercial deepfake detectors. We discuss tradeoffs in accuracy, computational costs, and interpretability, focusing on how practical these deepfake detectors might be when deployed in real-world settings.  Our code can be found in this \href{https://github.com/Deepfake-Detection-KKO/deepfake-detection}{repo}.

\end{abstract}

\section{Introduction}

\setlength{\textfloatsep}{10.0pt plus 6.0pt minus 6.0pt}

\begin{figure}
    \centering
    \caption{Real and fake image samples from Deepfake-Eval-2024}
    \includegraphics[width = 3.25in,trim={0.1in 0.1in 0.1in 0.1in},clip]{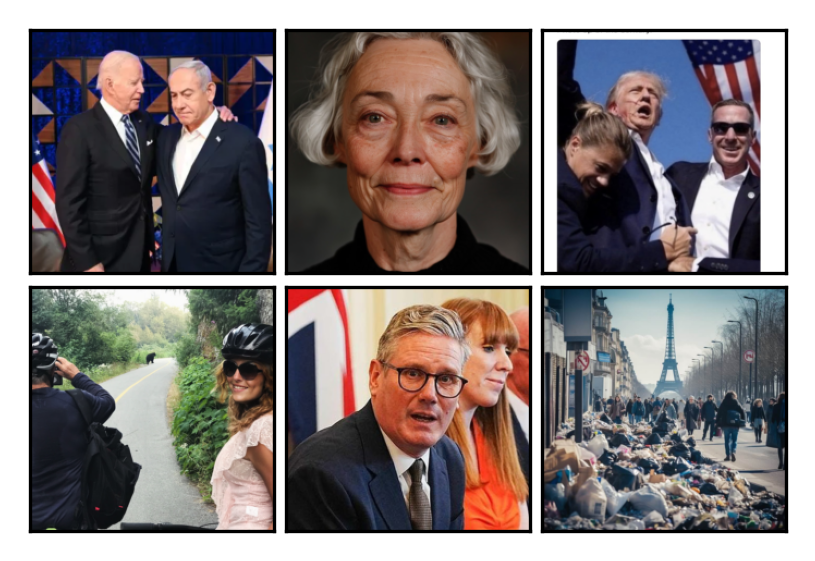}
    \label{gradcam_1}
\end{figure}

The proliferation of fake digital media presents ever-growing societal risks. Some individuals may use AI or other techniques to rapidly produce imagery with a malicious intent to disseminate disinformation, harass individuals, or commit fraud. Therefore, developing accurate deepfake detectors with real-world, high-quality benchmarks is critical for maintaining public trust and safety.

\textbf{Deepfake-Eval-2024}\cite{chandra_deepfake-eval-2024_2025} is a recently introduced benchmark for deepfake detection, focused on real-world, in-the-wild examples. This dataset contains 759 real images and 1,191 deepfake images collected from social media in 2024. The dataset represents a compelling distribution of recent ``in-the-wild'' deepfakes. 

The authors of Deepfake-Eval-2024 also reported the performance of three finetuned open-source models with accuracies ranging from 61\% to 69\%. In addition, the authors test several commercial deepfake detectors and report the highest accuracy -- 82\%. This suggests a significant gap remains between open-source models and the leading proprietary model. 

One of the simpler open-source models reported by Deepfake-Eval-2024 comes from previous work by Ojha et al.~\cite{ojha_towards_2023} which demonstrated that simple classifiers on top of the features of standard pretrained vision models yield deepfake detectors that generalize across different generative models. However, when initially evaluated on Deepfake-Eval-2024, this approach only achieves an accuracy of 63\%. 

We revisit this simple approach and demonstrate that it can actually perform much better -- 81\% accuracy on Deepfake-Eval-2024 -- outperforming the previously reported accuracy by 18\% and closely following the leading commerial model with 82\% accuracy. These results are achieved by optimizing several hyperparameters. Since these deepfake detectors perform nearly as well as commercial deepfake detectors, we consider their practical use in the real world and discuss the accuracy, computational cost, and interpretability tradeoffs offered by these models.

\section{Related Work}

Deepfake image detection has progressed from classical methods knowledge of manipulation techniques and bespoke feature extractions to modern deep learning methods which rely on large amounts of data\cite{durall2020unmaskingdeepfakessimplefeatures} \cite{mirsky_creation_and_detection_of_deepfakes_a_survey} \cite{heidari_deepfake_detection_using_dl_methods}. Up until recently, convolutional neural networks have been the dominant approach in deepfake detectors \cite{rana_deepfake_2022}  \cite{zhang_deepfake_2022}. The recent widespread adoption of new generative AI such as stable diffusion presents new challenges for deepfake detectors. At the same time, the introduction of vision transformers and next-generation convolutional neural networks offer new opportunities\cite{liu_convnet_2022} \cite{dosovitskiy_image_2021}. These evolutions demonstrate how deepfake generation and detection are intertwined \cite{mubarak_survey_on_the_detection_and_impacts_of_deepfakes} \cite{tassone_continuous_fake_media_detection}\cite{abbas_unmasking_deepfakes}.

Current deepfake detection research typically evaluates deepfake detectors on media generated by specific models or alteration methods \cite{ojha_towards_2023}. While valuable, the extent to which this work applies to the diversity of real-world deepfakes is limited. Real-world deepfakes span a range of subject matters, generative models, alteration methods, and sources \cite{zi_wilddeepfake} \cite{cho_towards_understanding_deepfake_videos_in_the_wild}. 

There also exists a growing body of research on ``in-the-wild'' deepfake detection. However, many of these works \cite{zi_wilddeepfake} \cite{cho_towards_understanding_deepfake_videos_in_the_wild} \cite{pu_deepfake_videos_in_the_wild} \cite{zhou_face_forensics_in_the_wild} \cite{liu_towards_spoofed_and_deepfake_speech_detection_in_the_wild} typically focus on detecting facial manipulations in human subjects as well as video or audio deepfakes , which represent an important but partial view of real-world deepfakes. Other works create deepfake datasets by altering untampered real-world images\cite{pirogov2025evaluatingdeepfakedetectorswild} or rely heavily on user-generated social media tagging\cite{cavia2024realtimedeepfakedetectionrealworld}\cite{batra2025socialdfbenchmarkdatasetdetection}, resulting in a deepfake distribution that may not fully capture the range of misinformation and harmful content.

Our work focuses on a new in-the-wild benchmark, Deepfake-Eval-2024. Originally introduced by Chandra et al.\cite{chandra_deepfake-eval-2024_2025}, it is one of the few in-the-wild deepfake benchmarks containing both real and fake images from social media. These samples were sourced directly from social media or by direct upload to TrueMedia.org, a deepfake detection platform.  The data was labeled by forensic analysts through a process incorporating reverse image search, evaluation of original sources, and inspection of typical AI-generated features. The dataset is diverse in its content and style, containing close-up portraits, large crowds, as well as non-human subjects. 

The authors of Deepfake-Eval-2024 also report the performance of three finetuned open-source models, ranging from 61\% to 69\% accuracy, along with the best tested commercial model, with 82\% accuracy. The initial reporting suggests a wide gap between open-source models and leading commercial deepfake detectors.  

One of the simpler open-source baselines was originally introduced by Ojha et al. \cite{ojha_towards_2023}. This work demonstrated that the surprisingly simple approach of training simple classifiers, like KNN or a single linear layers, on top of the feature space of pretrained vision models, like ViT-b16 or ResNet-50, yield deepfake detectors that generalize across generative models.  This is because they adapt the high-level semantic features from CLIP and ImageNet pretraining and do not rely on the highly-specific deepfake patterns from any particular generative model. However, these models were only tested on lab-generated deepfake images. As suggested by Chandra et al., this approach seems to fail accurate detection of real-world deepfakes, only achieving an accuracy of 63\% \cite{chandra_deepfake-eval-2024_2025}.

Our work revisits this baseline approach and extends the initial analysis by Deepfake-Eval-2024 in two ways. First, by further optimizing several hyperparameters with three different model architectures, we demonstrate how the simple approach of adapting standard pretrained vision models as deepfake detectors can actually achieve accuracy on par with the leading proprietary models on Deepfake-Eval-2024. Second, we discuss the tradeoffs between accuracy, model size, computational cost, and interpretability of these large open-source models. Deploying these deepfake detectors in real-world scenarios ultimately depends on how efficient and trustworthy they are.

\section{Approach}
\subsection{Experimental Design}
To train and evaluate deepfake detectors, we adhere to the same finetuning split prescribed by Deepfake-Eval-2024 which designates 60\% (1161 images) for training and 40\% (789 images) for testing. The original dataset can be found \href{https://huggingface.co/datasets/nuriachandra/Deepfake-Eval-2024/tree/main}{here}. From the training data, we further reserve 10\% (116 images) of training samples for validation during our hyperparameter search. This choice maximizes learning from more training samples at the cost of slightly greater uncertainty in accuracy from fewer validation samples. 

\subsection{Models}
\begin{figure}
    \centering
    \caption{Deepfake Detector Modeling Approach}
    \includegraphics[width = 3.25in]{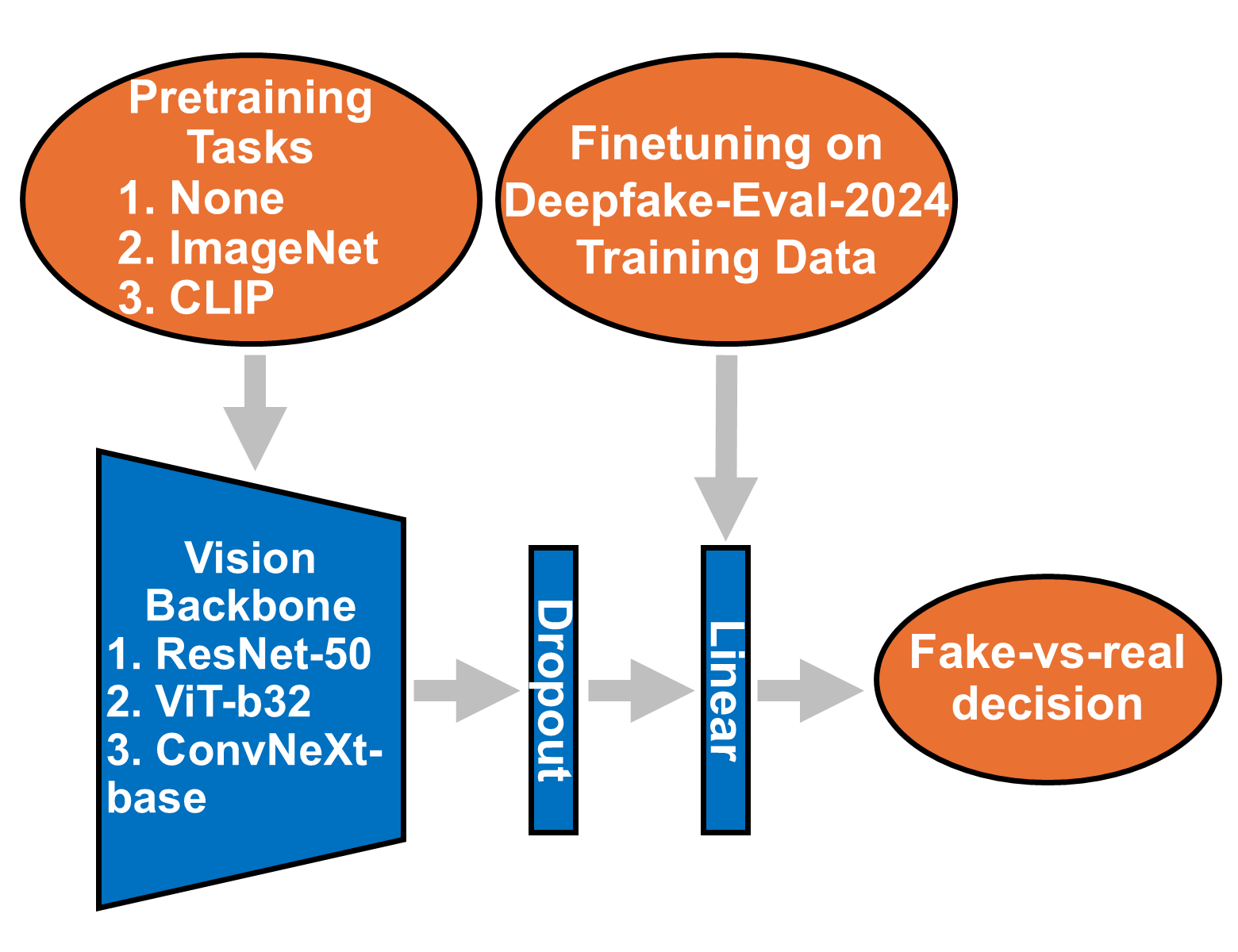}
    \label{deepfake_detector_architecture}
\end{figure}
Our deepfake detectors consist of a dropout layer and a single linear layer on top of three standard vision models: \textbf{ResNet-50}, \textbf{ViT-b32}, and \textbf{ConvNeXt-base}. These are standard variants within each architecture family. ResNet-50 is a classic convolutional neural network and baseline from 2015\cite{he_deep_2015}. ViT-b32 adapts transformers, popular in NLP, for computer vision\cite{dosovitskiy_image_2021}. ConvNeXt represents the latest generation of CNNs by incorporating advancements from transformers such as larger effective kernels, GELU activations, and batch norm into a pure convolutional approach\cite{liu_convnet_2022}. Previous work has established that different model architectures identify distinct types of anomalies, artifacts, and patterns for deepfake detection\cite{afchar_mesonet_2018}. Our choice of ResNet-50, ViT-b32, and ConvNeXt-base highlights these distinct capabilities. 

We use randomly-initialized weights as well as learned weights from ImageNet and CLIP pretraining. Models trained from scratch serve as a baseline to quantify the marginal benefit of the learned features from ImageNet and CLIP. With pretrained models, we experimented with both freezing the backbone as well as finetuning all weights in the network to understand how well the learned features transfer to deepfake detection without modification. Figure \ref{deepfake_detector_architecture} illustrates the overall modeling approach.

\subsection{Data Processing}
We implement several procedures to accelerate the training process. The smaller dimension of all training images is rescaled to 384 pixels. This resolution speeds up the data transfer and processing compared to preliminary experiments, while maintaining enough detail for our models to learn from. 

At training time, images are first randomly cropped between 50\% and 100\% of the original size, then further rescaled to 256 pixels in order to augment the training data. At validation and test time, no cropping or rescaling was applied. While using pretrained weights, we also apply the same preprocessing used during the pretraining tasks. This primarily means that images are normalized according to the distribution of ImageNet or CLIP. 

\subsection{Training}
We utilize Adam optimization algorithm since it typically converges quickly with minimal tuning. We implement early stopping after training for 5 epochs without improvement in validation performance, which helps avoid overfitting and speeds up our experiments.

To modulate the learning rate, we test both step decay scheduling and cosine annealing. We configure our step scheduler to halve the learning rate every two epochs. Cosine annealing cycles between learning rate decay and warm restarts i.e. resetting the learning rate to a higher level.

Finally, since the distribution of fake versus real images in Deepfake-Eval-2024 is not skewed enough to warrant a class-balanced loss or re-weighting scheme, we opted to use standard cross-entropy loss. 

\section{Experiment and Results}
\subsection{Evaluation Metrics}
We primarily used accuracy because it is a simple metric and allows direct comparison of our modeling results with those reported in Deepfake-Eval-2024\cite{chandra_deepfake-eval-2024_2025}. However, there's no single consistently used metric for evaluating the predictive capability of deepfake detectors. In the real world, deepfake detectors might be deployed in different ways for use cases such as social media content moderation or threat intelligence. To this end, we consider ROC AUC and mean average precision which summarize detection rates, false positive rates, and precision at different operating points.  

In addition to measuring predictive performance, we also consider the time and space complexity of our models when discussing the performance of our final tuned models. We examined these models' size, floating point operations, and inference speed. These metrics have implications for how reliable deepfake detectors are in the real world.

\subsection{Hyperparameter Tuning Results}
\begin{figure*}
    \caption{ConvNeXt-base Learning Curves}
    \centering
    \includegraphics[trim={0 0.3in 0 0.1in},clip]{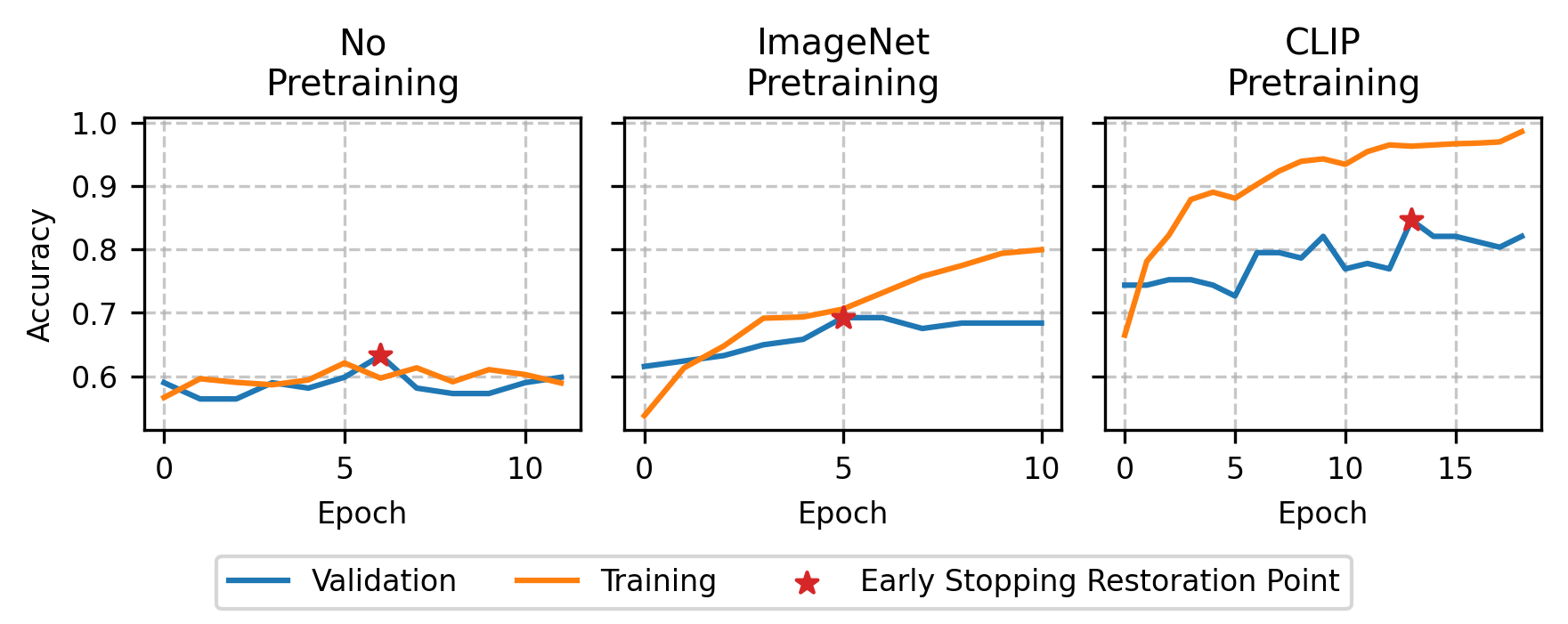}
    \includegraphics[trim={0 0 0 0.05in},clip]{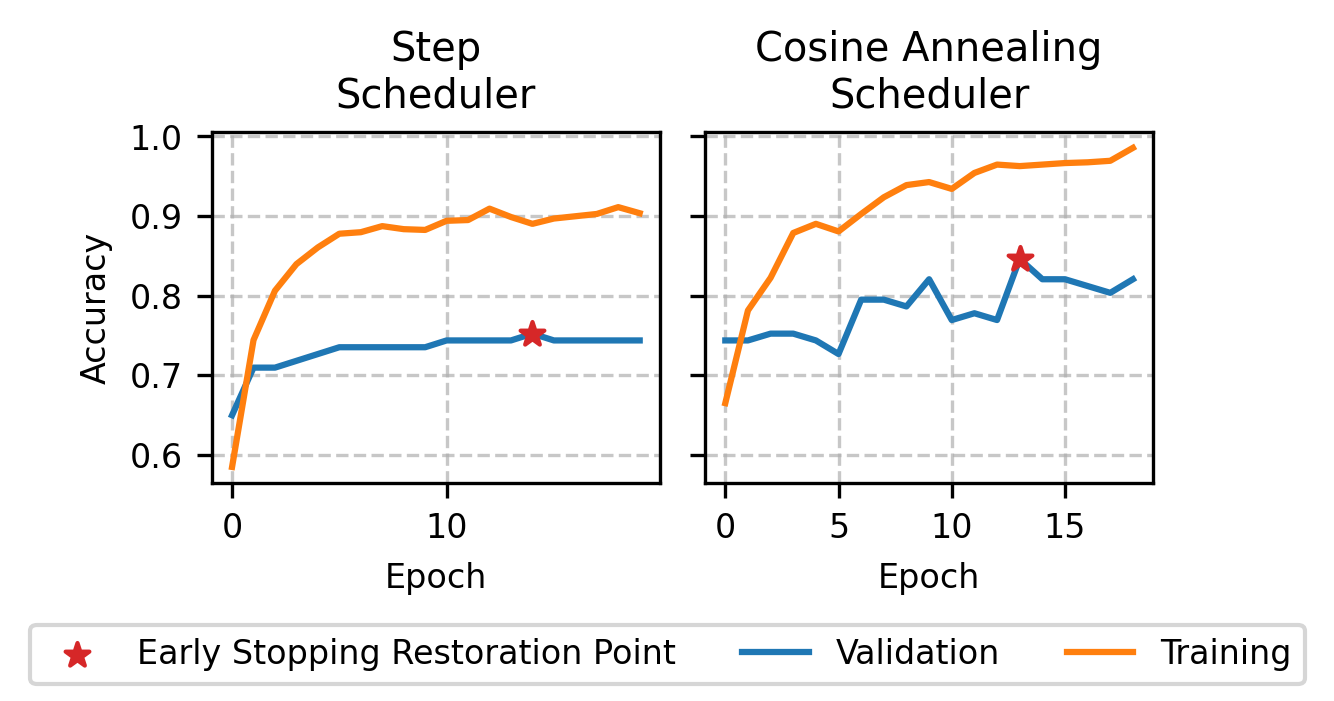}
    \label{learning_curves_by_scheduler}
\end{figure*}

Our preliminary experiments demonstrate that freezing the pretrained backbones while only training the linear classification head results in worse performance than finetuning the entire network's weights. This suggests that the extracted features from ImageNet or CLIP, which transfer well to many other image recognition and object detection tasks, are suboptimal for identifying the subtle differences between deepfakes and real images without additional finetuning. Freezing the backbone's weights is also not conducive to visualizing gradient-weighted \textit{fake-vs-real} activation mappings. This visualization depends on how the internal \textit{trained} parts of the networks learn to recognize deepfakes. We discuss these visualizations in a following section. For these reasons, we prioritize finetuning all weights of our models.

Without pretraining, we initially experimented with a wider range of learning rates and longer training duration with the expectation that models would learn at least a small amount. However, their training loss hardly drops as illustrated by the top left plot of figure \ref{learning_curves_by_scheduler}, suggesting either poor initialization or vanishing gradients. The Deepfake-Eval-2024 training dataset is likely too small to train a model with random initializations. Our preliminary experiments also indicate that most models benefit from some moderate regularization in the form of L2 penalty and dropout preceding the final classification. More punitive L2 penalties or dropout rates did not significantly improve generalization. Therefore, we prioritize tuning these parameters at only moderate levels.

\begin{table*}
\caption{Best Validation Accuracies from Hyperparameter Tuning}
\label{tab:validation_results}
\begin{adjustbox}{width=1.0\textwidth}
\setlength\tabcolsep{3pt}
\begin{tabular}{l|l||ccc|ccc|ccc}
\toprule
 \cellcolor[HTML]{FFFFFF} & Model Architecture & \multicolumn{3}{c|}{ResNet-50} & \multicolumn{3}{c|}{ViT-b32} & \multicolumn{3}{c}{ConvNeXt-base} \\
 \hline
 
 \cellcolor[HTML]{FFFFFF} & \multirow[c]{3}{*}{\diagbox[width=1.3in, innerwidth=1.1in, height=3\line]{\\Initial\\Learning Rate}{Pretraining\\Task}} & \cellcolor[HTML]{FFFFFF} & \cellcolor[HTML]{FFFFFF} & \cellcolor[HTML]{FFFFFF} & \cellcolor[HTML]{FFFFFF} & \cellcolor[HTML]{FFFFFF} & \cellcolor[HTML]{FFFFFF} & \cellcolor[HTML]{FFFFFF} & \cellcolor[HTML]{FFFFFF} & \cellcolor[HTML]{FFFFFF} \\

& & & & & & & & & & \\ 
 
  \multirow[c]{-3}{*}{\cellcolor[HTML]{FFFFFF}\parbox{1in}{Learning Rate\\Scheduler}} & & \multirow[c]{-3}{*}{\cellcolor[HTML]{FFFFFF}\parbox{0.5in}{\centering None}} & \multirow[c]{-3}{*}{\cellcolor[HTML]{FFFFFF}\parbox{0.5in}{\centering ImageNet}} & \multirow[c]{-3}{*}{\cellcolor[HTML]{FFFFFF}\parbox{0.5in}{\centering CLIP}} & \multirow[c]{-3}{*}{\cellcolor[HTML]{FFFFFF}\parbox{0.5in}{\centering None}} & \multirow[c]{-3}{*}{\cellcolor[HTML]{FFFFFF}\parbox{0.5in}{\centering ImageNet}} & \multirow[c]{-3}{*}{\cellcolor[HTML]{FFFFFF}\parbox{0.5in}{\centering CLIP}} & \multirow[c]{-3}{*}{\cellcolor[HTML]{FFFFFF}\parbox{0.5in}{\centering \hspace{-0.5pt}None}} & \multirow[c]{-3}{*}{\cellcolor[HTML]{FFFFFF}\parbox{0.55in}{\centering \hspace{-0.5pt}ImageNet}} & \multirow[c]{-3}{*}{\cellcolor[HTML]{FFFFFF}\parbox{0.5in}{\centering \hspace{-0.5pt}CLIP}} \\
\hline
\hline

\cellcolor[HTML]{FFFFFF} & 1e-3 & {\cellcolor[HTML]{89BEDC}} \color[HTML]{000000} 0.709 & {\cellcolor[HTML]{135FA7}} \color[HTML]{F1F1F1} 0.803 & {\cellcolor[HTML]{4896C8}} \color[HTML]{F1F1F1} 0.752 & {\cellcolor[HTML]{E9F2FA}} \color[HTML]{000000} 0.624 & {\cellcolor[HTML]{DBE9F6}} \color[HTML]{000000} 0.641 & {\cellcolor[HTML]{F7FBFF}} \color[HTML]{000000} 0.607 & {\cellcolor[HTML]{E3EEF8}} \color[HTML]{000000} 0.632 & {\cellcolor[HTML]{2171B5}} \color[HTML]{F1F1F1} 0.786 & {\cellcolor[HTML]{DBE9F6}} \color[HTML]{000000} 0.641 \\

\cellcolor[HTML]{FFFFFF} & 1e-4 & {\cellcolor[HTML]{BCD7EB}} \color[HTML]{000000} 0.675 & {\cellcolor[HTML]{0B559F}} \color[HTML]{F1F1F1} 0.812 & {\cellcolor[HTML]{2171B5}} \color[HTML]{F1F1F1} 0.786 & {\cellcolor[HTML]{DBE9F6}} \color[HTML]{000000} {\cellcolor[HTML]{DBE9F6}} \color[HTML]{000000} {\cellcolor[HTML]{DBE9F6}} \color[HTML]{000000} 0.641 & {\cellcolor[HTML]{3484BF}} \color[HTML]{F1F1F1} 0.769 & {\cellcolor[HTML]{DBE9F6}} \color[HTML]{000000} 0.641 & {\cellcolor[HTML]{F1F7FD}} \color[HTML]{000000} 0.615 & {\cellcolor[HTML]{135FA7}} \color[HTML]{F1F1F1} 0.803 & {\cellcolor[HTML]{084B93}} \color[HTML]{F1F1F1} 0.821 \\

\multirow[c]{-3}{*}{\cellcolor[HTML]{FFFFFF}\parbox{1in}{Cosine Annealing}} & 1e-5 & {\cellcolor[HTML]{CDE0F1}} \color[HTML]{000000} 0.658 & {\cellcolor[HTML]{1967AD}} \color[HTML]{F1F1F1} 0.795 & {\cellcolor[HTML]{2A7AB9}} \color[HTML]{F1F1F1} 0.778 & {\cellcolor[HTML]{A4CCE3}} \color[HTML]{000000} {\cellcolor[HTML]{A4CCE3}} 0.692 & {\cellcolor[HTML]{084B93}} \color[HTML]{F1F1F1} 0.821 & {\cellcolor[HTML]{083877}} \color[HTML]{F1F1F1} 0.838 & {\cellcolor[HTML]{CDE0F1}} \color[HTML]{000000} 0.658 & {\cellcolor[HTML]{5FA6D1}} \color[HTML]{F1F1F1} 0.735 & {\cellcolor[HTML]{08306B}} \color[HTML]{F1F1F1} 0.846 \\
\hline

\cellcolor[HTML]{FFFFFF} & 1e-3 & {\cellcolor[HTML]{1967AD}} \color[HTML]{F1F1F1} 0.795 & {\cellcolor[HTML]{135FA7}} \color[HTML]{F1F1F1} 0.803 & {\cellcolor[HTML]{539ECD}} \color[HTML]{F1F1F1} 0.744 & {\cellcolor[HTML]{F7FBFF}} \color[HTML]{000000} 0.607 & {\cellcolor[HTML]{D3E4F3}} \color[HTML]{000000} 0.65 & {\cellcolor[HTML]{F7FBFF}} \color[HTML]{000000} 0.607 & {\cellcolor[HTML]{E3EEF8}} \color[HTML]{000000} 0.632 & {\cellcolor[HTML]{2A7AB9}} \color[HTML]{F1F1F1} 0.778 & {\cellcolor[HTML]{F7FBFF}} \color[HTML]{000000} 0.607 \\

 \cellcolor[HTML]{FFFFFF}& 1e-4 & {\cellcolor[HTML]{BCD7EB}} \color[HTML]{000000} 0.675 & {\cellcolor[HTML]{135FA7}} \color[HTML]{F1F1F1} 0.803 & {\cellcolor[HTML]{0B559F}} \color[HTML]{F1F1F1} 0.812 & {\cellcolor[HTML]{DBE9F6}} \color[HTML]{000000} 0.641 & {\cellcolor[HTML]{A4CCE3}} \color[HTML]{000000} 0.692 & {\cellcolor[HTML]{DBE9F6}} \color[HTML]{000000} 0.641 & {\cellcolor[HTML]{E9F2FA}} \color[HTML]{000000} 0.624 & {\cellcolor[HTML]{2A7AB9}} \color[HTML]{F1F1F1} 0.778 & {\cellcolor[HTML]{083877}} \color[HTML]{F1F1F1} 0.838 \\
 
 \multirow[c]{-3}{*}{\cellcolor[HTML]{FFFFFF}\parbox{1in}{Step Decay}}& 1e-5 & {\cellcolor[HTML]{DBE9F6}} \color[HTML]{000000} 0.641 & {\cellcolor[HTML]{89BEDC}} \color[HTML]{000000} 0.709 & {\cellcolor[HTML]{6CAED6}} \color[HTML]{F1F1F1} 0.726 & {\cellcolor[HTML]{C6DBEF}} \color[HTML]{000000} 0.667 & {\cellcolor[HTML]{5FA6D1}} \color[HTML]{F1F1F1} 0.735 & {\cellcolor[HTML]{0B559F}} \color[HTML]{F1F1F1} 0.812 & {\cellcolor[HTML]{F1F7FD}} \color[HTML]{000000} 0.615 & {\cellcolor[HTML]{7AB6D9}} \color[HTML]{000000} 0.718 & {\cellcolor[HTML]{1967AD}} \color[HTML]{F1F1F1} 0.795 \\

\bottomrule
\end{tabular}
\end{adjustbox}
\end{table*}

Table \ref{tab:validation_results} summarizes our most significant experimental results. The table shows both the initial learning rate and scheduler make a substantial difference on accuracy, particularly with the larger ViT-b32 and ConvNeXt-base models. Generally, a lower learning rate and cosine annealing lead to higher accuracy than step decay. Cosine annealing involves gradually decreasing the learning rate following a cosine curve. The learning rate periodically resets to a higher level, helping the model escape local minima and explore more of the loss landscape. Later in the training process, the slower learning rate helps the model navigate more precisely toward better optima. The smaller gap between training and validation loss in Figure \ref{learning_curves_by_scheduler} also demonstrate how cosine annealing's warm restarts mitigate some of the step scheduler's overfitting. These results underscore how sustained exploration induced by annealing leads to our most generalizable deepfake detectors. 

Table \ref{tab:validation_results} also shows the signficant impact that larger pretrained vision models have on accuracy. Unsurprisingly, pretraining produces models with accuracy exceeding models with randomly-initialized weights by 10-20\%. ConvNeXt-base pretrained on CLIP performs the best indicating that this architecture's representational capacity and learned features from CLIP work particularly well together for deepfake detection. Since the ConvNeXt architecture is a CNN that utilizes larger kernel sizes for more global receptive fields, it's inductive biases could make it well-suited for detecting both low-level artifacts as well as high-level features that occur in deepfakes. Furthermore, contrastive language-image pretraining on LAION-400M\cite{githubOpen_clipdocsmodel_profilecsvMain}\cite{Cherti_2023}, a large-scale image-text dataset from 2021, mostly likely exposed the model to some number of deepfakes and more recent images from the internet. This contrasts with ImageNet, a smaller dataset of images from 2010, which are less representative of social media-sourced images in the Deepfake-Eval-2024 benchmark. With early stopping, CLIP pretraining results in less overfitting and more generalizable deepfake detectors as illustrated by the top learning curves in Figure \ref{learning_curves_by_scheduler}.

ViT-b32 performs nearly as well with CLIP embeddings likely due to a large learning capacity comparable to ConvNeXt. ViT-b32's larger effective receptive field suits it detecting more global deepfake patterns in the images. ResNet-50's best accuracy is lower, likely due to to the smaller effective receptive fields associated with early layers of ResNet-50 compared to ViT-b32 \cite{raghu_vision_2022} or ConvNeXt. Furthermore, ResNet-50 has the fewest parameters and smallest architecture, making it the least expressive model.

\subsection{Test Performance Results}
\begin{figure}
    \centering
    \includegraphics[width=1\linewidth]{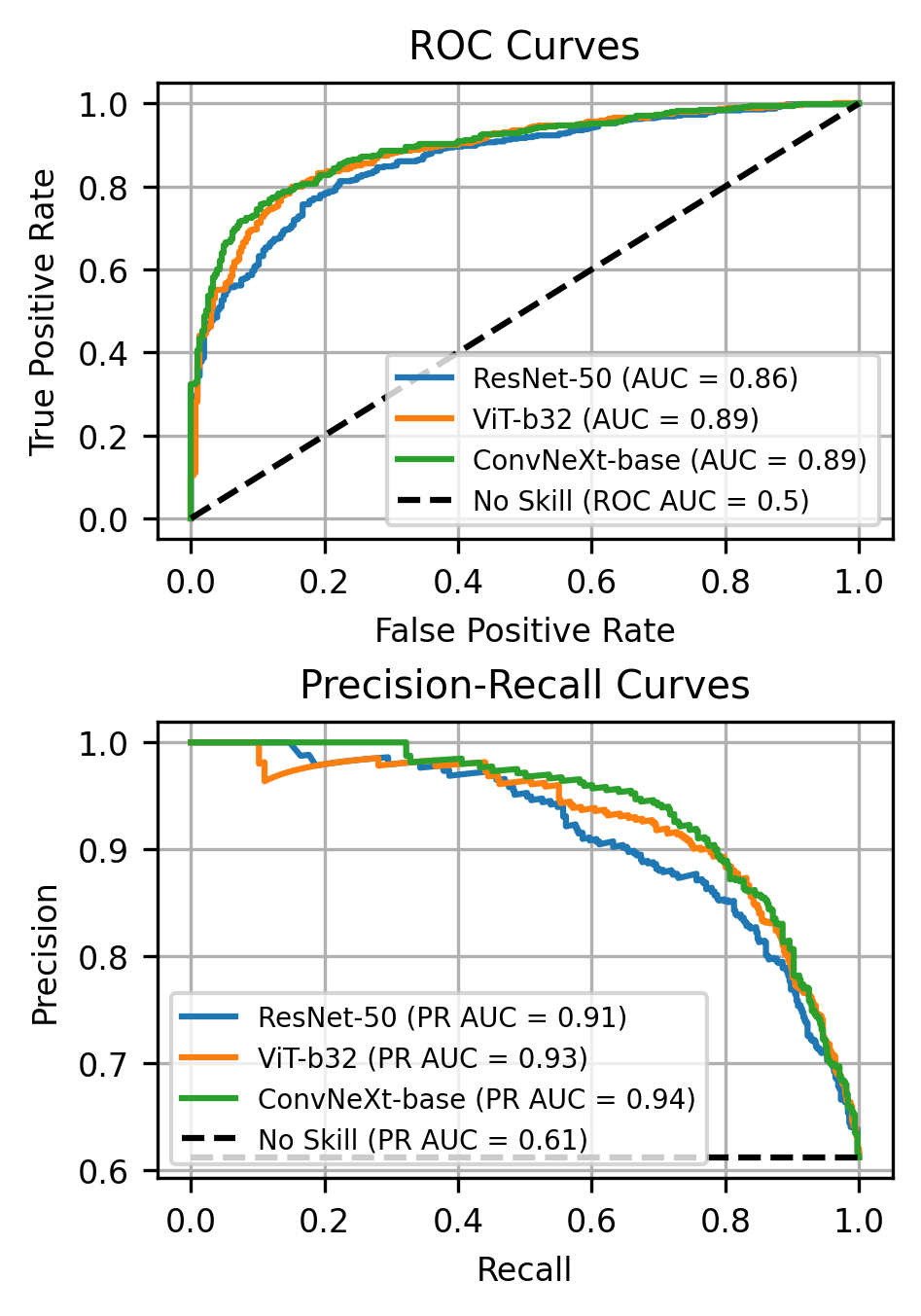}
    \caption{Performance on Test Data}
    \label{roc_pr_curves}
\end{figure}
\begin{table*}
\caption{Test Accuracy and Computational Performance}
\label{tab:test_accuracy_computational_performance}
\begin{center}
\begin{tabular}{r||cl|cl|cl|cl}
\toprule
Model & \multicolumn{2}{c}{\centering Test Accuracy} & \multicolumn{2}{p{2cm}}{\centering Parameters (millions)\cite{githubOpen_clipdocsmodel_profilecsvMain}} & \multicolumn{2}{c}{\centering GFLOPs\cite{githubOpen_clipdocsmodel_profilecsvMain}} & \multicolumn{2}{p{2cm}}{\centering Average Inference Speed (ms)*} \\
\midrule
ResNet-50 & 0.79 & \tablebar{0.96}{teal} & 38.32 & \tablebar{0.44}{purple} & 12.22 & \tablebar{0.4}{violet} & 21.96 & \tablebar{0.5}{violet} \\
ViT-b32 & 0.81 & \tablebar{0.98}{teal} & 87.85 & \tablebar{1.0}{purple} & 8.82 & \tablebar{0.29}{violet} & 16.0 & \tablebar{0.36}{violet} \\
ConvNeXt-base & 0.81 & \tablebar{0.99}{teal} & 88.09 & \tablebar {1.0}{purple} & 30.71 & \tablebar{1.0}{violet} & 44.25 & \tablebar{1.0}{violet} \\
Previous Best Open-Source Model\cite{chandra_deepfake-eval-2024_2025} & 0.69 & \tablebar{0.84}{teal} & \multicolumn{2}{c}{\centering -- } & \multicolumn{2}{c}{\centering -- } & \multicolumn{2}{c}{\centering -- } \\
Leading Commercial Model\cite{chandra_deepfake-eval-2024_2025} & 0.82 & \tablebar{1.0}{teal} & \multicolumn{2}{c}{\centering -- } & \multicolumn{2}{c}{\centering -- } & \multicolumn{2}{c}{\centering -- } \\
No Skill Classifier** & 0.61 & \tablebar{0.74}{teal} & \multicolumn{2}{c}{\centering -- } & \multicolumn{2}{c}{\centering -- } & \multicolumn{2}{c}{\centering -- } \\
\bottomrule
\end{tabular}
\end{center}
\vspace{8pt}
* Tested on H200 GPUs with batch size of 1\\
** Represents always choosing the majority class from the training set (``fake")
\end{table*}

Figure \ref{roc_pr_curves} shows the precision, detection, and false positive rates of our most accurate ConvNeXt-base, ResNet-50, and ViT-b32 models on the test dataset. ConvNeXt performs slightly better at most operating points which are summarized by its higher ROC AUC values of 0.89 and mean precision of 0.94. The precision-recall curve specifically shows that our ConvNeXt-base model allows detection of more than 30\% of deepfakes with zero false positives on the Deepfake-Eval-2024 benchmark. This suggests how our ConvNeXt-base model could be used in industry settings where precision is critical -- filtering out a significant portion of deepfakes without the need for manual review by content reviewers or analysts.

Table \ref{tab:test_accuracy_computational_performance} shows our same deepfake detectors' accuracy along with measurements of inference speed, floating point operations (GFLOPs), and the number of model parameters. This table establishes ConvNeXt-base and ViT-b32 as our most accurate models for the Deepfake-Eval-2024 benchmark. However, ViT-b32 is nearly three times as fast as measured by our tests and reported floating point operations per prediction\cite{githubOpen_clipdocsmodel_profilecsvMain}. This highlights a distinct advantage that ViT-b32's architecture has over ConvNeXt-base. ViT-b32 immediately downsamples the input image into a few large 32x32 patches, resulting in fewer computations from early on. On the other hand, ConvNeXt-base gradually downsamples using convolutions with small strides which require many more computations throughout the network. This suggests that ViT-b32 may be preferred to ConvNeXt-base when predictions need to be being served for time-sensitive tasks or in high-throughput workloads. ResNet-50 slightly lags ViT-b32 and ConvNeXt-base with an accuracy of 79\%. Because ResNet-50's size is much smaller, it may be preferred to deployed on smaller computers or devices where memory is more limited.

Table \ref{tab:test_accuracy_computational_performance} also compares the accuracy of our models with previously-reported models on Deepfake-Eval-2024\cite{chandra_deepfake-eval-2024_2025}. All three of our models surpass the accuracy of the previous leading open-source models by 10-12\%. Furthermore, Table \ref{tab:test_accuracy_computational_performance} shows that large open-source vision models can be trained as deepfake detectors which closely compete with the leading commercial models at around 82\% accuracy. Our results show how the approach of pretraining on CLIP and further finetuning slowly with cosine annealing sets a new standard for open-source models on Deepfake-Eval-2024.

\subsection{Error Analysis \& Visualization}
\begin{figure}
    \centering
    \includegraphics[width = 3.15in,trim={0.05in 0.075in 0.1in 0.1in},clip]{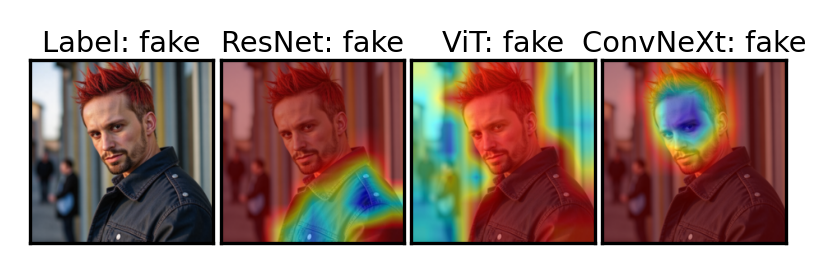}
    \\[0.05cm]
    \includegraphics[width = 3.15in,trim={0.05in 0.075in 0.1in 0.1in},clip]{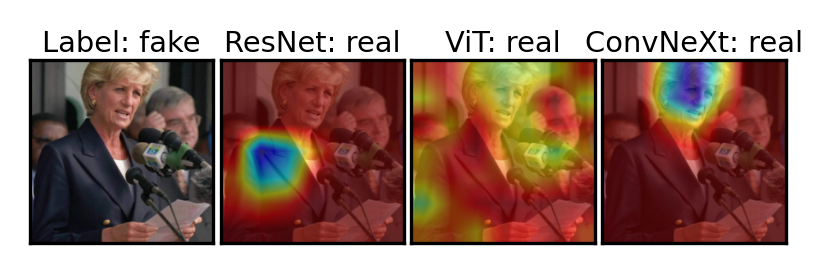}
    \\[0.05cm]
    \includegraphics[width = 3.15in,trim={0.05in 0.075in 0.1in 0.1in},clip]{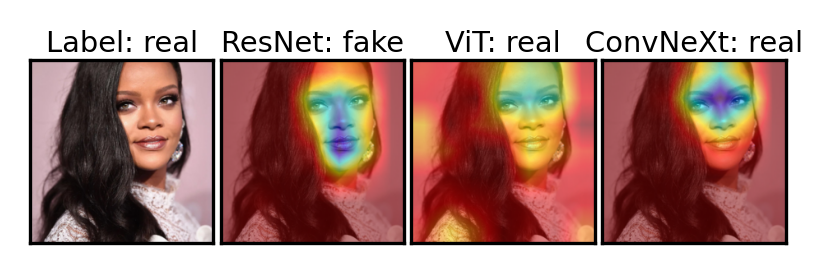}
    \\[0.05cm]
    \includegraphics[width = 3.15in,trim={0.05in 0.075in 0.1in 0.1in},clip]{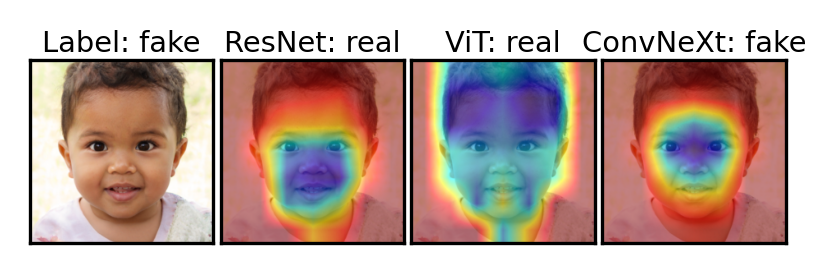}
    \\[0.05cm]
    \includegraphics[width = 3.15in,trim={0.05in 0.075in 0.1in 0.1in},clip]{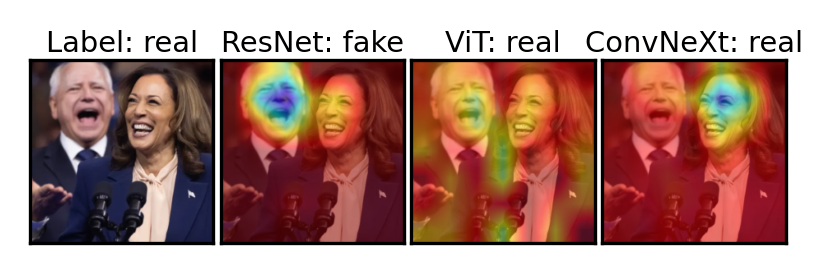}
    \\[0.05cm]
    \includegraphics[width = 3.15in,trim={0.05in 0.075in 0.1in 0.1in},clip]{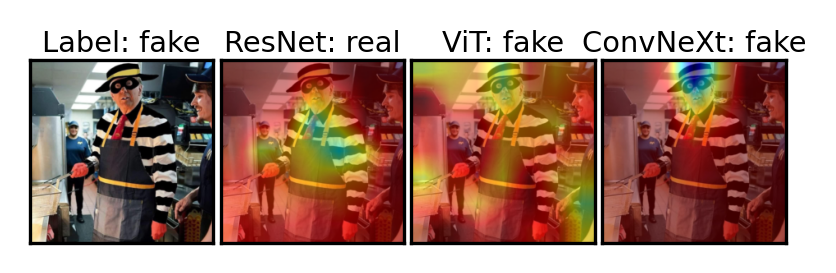}
    \\[0.05cm]
    \caption{GradCAM Visualizations (important regions in blue)}
    \label{gradcam_1}
\end{figure}

End-users may scrutinize the trustworthiness of deepfake detectors based on large deep learning models. While testing on realistic benchmarks such as Deepfake-Eval-2024 demonstrate their reliability, we can identify reoccurring points of failure, dataset biases, and further improve our deepfake detectors by reviewing test samples and understanding of how the models learn to ``see'' them.

Figure \ref{gradcam_1} shows gradient-weighted class activation mapping (GradCAM) visualizations on six test samples from Deepfake-Eval-2024. GradCAM is a technique for visualizing how our vision models make fake-vs-real decisions. It depends on combining feature maps from one of the last internal model layers. Before combining these feature maps, they are weighed by the gradient of the predicted class output with respect to this last layer. They are combined and ReLU-activated to produce a heatmap highlighting regions our models focus on when deciding fake-vs-real. 

The six images from Figure \ref{gradcam_1} were selected to illustrate the most common observations from our review: 
\begin{itemize}
    \item \textbf{All models focused on different visual cues and regions of an image even when they agree.} This is no surprise as each model's distinct architecture will naturally learn to produce different feature maps. This is most clearly illustrated by the first and second row in Figure \ref{gradcam_1}.
    \item \textbf{ViT-b32 more often highlighted larger and several disparate parts of an image whereas ConvNeXt-base and ResNet-50 focused on one smaller region.} This directly relates to the self-attention mechanism in the transformer architecture in which image patches connect to each other, effectively learning more global image patterns. On the other hand, convolutional neural networks analyze patches in a localized manner, gradually building a hierarchical set of features. This is well-established and analyzed by previous deepfake detection and computer vision research and underscore how CNNs and transformers may compliment each other for deepfakes with both low- and high-level features. \cite{wodajo_deepfake_2021}  \cite{afchar_mesonet_2018} \cite{raghu_vision_2022}. This pattern appears in most images of Figure \ref{gradcam_1}.
    \item \textbf{ResNet-50 and ConvNeXt-base frequently focused on different facial features.} ResNet-50 more often erred by focusing on lower facial features such as the nose and mouth. On the other hand, ConvNeXt-base looked at upper facial features such as the eyes and nasal bridge. It's possible some particular face swap or alteration techniques are more prevalent Deepfake-Eval-2024. ConvNeXt's higher learning capacity may allow it to identify such manipulations more effectively than the smaller ResNet-50 model. This is best illustrated in the third, fourth, and fifth row in Figure \ref{gradcam_1}.
    \item \textbf{Not all fake images were AI-generated.} Some of the fake images found in Deepfake-Eval-2024 seem to be produced using more primitive techniques or software. This highlights the unique labeling methodology of Deepfake-Eval-2024. TrueMedia.org reviewers often labeled the ground truth of these image based on the trustworthiness of the original source or if the photo appears to be edited in any manner. This calls attention to whether developing catch-all `deepfake' detectors is the most effective approach.  The last row in Figure \ref{gradcam_1} appears be a real photo modified with widely available image-editing software. Only ViT-b32 decides this image was fake.
\end{itemize}

\section{Conclusion}

We've demonstrated the simple baseline of finetuning standard pretrained vision-language models on Deepfake-Eval-2024 yields an accuracy of 81\% -- far surpassing this approach's initially reported 63\% accuracy and competing with the leading proprietary deepfake detector, with an accuracy of 82\%. Our best deepfake detectors used ViT-b32 and ConvNeXt-base architectures, pretrained on CLIP, and finetuned further with a low learning rate and cosine annealing. We've also highlighted some tradeoffs these open-source models offer in terms of size, complexity, and interpretability. 

We acknowledge the great progress Deepfake-Eval-2024 makes as a new in-the-wild deepfake detection benchmark. At the same time, it is clear its small size and distribution are challenging. While the particular collection channels and labeling methodology used to curate this dataset make our benchmark results noteworthy, they may not necessarily represent how these deepfake detectors perform across all real-world settings. 

Furthermore, the field of artificial intelligence and deepfake generation techniques is advancing quickly. It is critical to continue dedicating resources to developing newer and realistic benchmarks as well as new models to keep up with these rapid advancements. Deepfake detection will continue to matter for maintaining the public's vigilance of disinformation, fraud, and harassment as well as trust in digital media.

{\small
\bibliographystyle{ieee_fullname}
\bibliography{egpaper_for_review}

\begin{thebibliography}{10}\itemsep=-1pt

\bibitem{abbas_unmasking_deepfakes}
Fakhar Abbas and Araz Taeihagh.
\newblock Unmasking deepfakes: A systematic review of deepfake detection and generation techniques using artificial intelligence.
\newblock {\em Expert Systems with Applications}, 252:124260, 2024.

\bibitem{afchar_mesonet_2018}
Darius Afchar, Vincent Nozick, Junichi Yamagishi, and Isao Echizen.
\newblock {MesoNet}: a {Compact} {Facial} {Video} {Forgery} {Detection} {Network}.
\newblock pages 1--7, Dec. 2018.
\newblock arXiv:1809.00888 [cs].

\bibitem{batra2025socialdfbenchmarkdatasetdetection}
Arnesh Batra, Anushk Kumar, Jashn Khemani, Arush Gumber, Arhan Jain, and Somil Gupta.
\newblock Socialdf: Benchmark dataset and detection model for mitigating harmful deepfake content on social media platforms, 2025.

\bibitem{cavia2024realtimedeepfakedetectionrealworld}
Bar Cavia, Eliahu Horwitz, Tal Reiss, and Yedid Hoshen.
\newblock Real-time deepfake detection in the real-world, 2024.

\bibitem{chandra_deepfake-eval-2024_2025}
Nuria~Alina Chandra, Ryan Murtfeldt, Lin Qiu, Arnab Karmakar, Hannah Lee, Emmanuel Tanumihardja, Kevin Farhat, Ben Caffee, Sejin Paik, Changyeon Lee, Jongwook Choi, Aerin Kim, and Oren Etzioni.
\newblock Deepfake-{Eval}-2024: {A} {Multi}-{Modal} {In}-the-{Wild} {Benchmark} of {Deepfakes} {Circulated} in 2024, May 2025.
\newblock arXiv:2503.02857 [cs].

\bibitem{Cherti_2023}
Mehdi Cherti, Romain Beaumont, Ross Wightman, Mitchell Wortsman, Gabriel Ilharco, Cade Gordon, Christoph Schuhmann, Ludwig Schmidt, and Jenia Jitsev.
\newblock Reproducible scaling laws for contrastive language-image learning.
\newblock In {\em 2023 IEEE/CVF Conference on Computer Vision and Pattern Recognition (CVPR)}, page 2818–2829. IEEE, June 2023.

\bibitem{cho_towards_understanding_deepfake_videos_in_the_wild}
Beomsang Cho, Binh~M. Le, Jiwon Kim, Simon Woo, Shahroz Tariq, Alsharif Abuadbba, and Kristen Moore.
\newblock Towards understanding of deepfake videos in the wild.
\newblock In {\em Proceedings of the 32nd ACM International Conference on Information and Knowledge Management}, CIKM '23, page 4530–4537, New York, NY, USA, 2023. Association for Computing Machinery.

\bibitem{dosovitskiy_image_2021}
Alexey Dosovitskiy, Lucas Beyer, Alexander Kolesnikov, Dirk Weissenborn, Xiaohua Zhai, Thomas Unterthiner, Mostafa Dehghani, Matthias Minderer, Georg Heigold, Sylvain Gelly, Jakob Uszkoreit, and Neil Houlsby.
\newblock An {Image} is {Worth} 16x16 {Words}: {Transformers} for {Image} {Recognition} at {Scale}, June 2021.
\newblock arXiv:2010.11929 [cs].

\bibitem{durall2020unmaskingdeepfakessimplefeatures}
Ricard Durall, Margret Keuper, Franz-Josef Pfreundt, and Janis Keuper.
\newblock Unmasking deepfakes with simple features, 2020.

\bibitem{githubOpen_clipdocsmodel_profilecsvMain}
ML Foundations.
\newblock {OpenCLIP Model Profiles}.
\newblock \url{https://github.com/mlfoundations/open_clip/blob/main/docs/model_profile.csv}, 2023.
\newblock [Accessed 25-07-2025].

\bibitem{he_deep_2015}
Kaiming He, Xiangyu Zhang, Shaoqing Ren, and Jian Sun.
\newblock Deep {Residual} {Learning} for {Image} {Recognition}, Dec. 2015.
\newblock arXiv:1512.03385 [cs].

\bibitem{heidari_deepfake_detection_using_dl_methods}
Arash Heidari, Nima Jafari~Navimipour, Hasan Dag, and Mehmet Unal.
\newblock Deepfake detection using deep learning methods: A systematic and comprehensive review.
\newblock {\em WIREs Data Mining and Knowledge Discovery}, 14(2):e1520, 2024.

\bibitem{liu_towards_spoofed_and_deepfake_speech_detection_in_the_wild}
Xuechen Liu, Xin Wang, Md Sahidullah, Jose Patino, Héctor Delgado, Tomi Kinnunen, Massimiliano Todisco, Junichi Yamagishi, Nicholas Evans, Andreas Nautsch, and Kong~Aik Lee.
\newblock Asvspoof 2021: Towards spoofed and deepfake speech detection in the wild.
\newblock {\em IEEE/ACM Transactions on Audio, Speech, and Language Processing}, 31:2507--2522, 2023.

\bibitem{liu_convnet_2022}
Zhuang Liu, Hanzi Mao, Chao-Yuan Wu, Christoph Feichtenhofer, Trevor Darrell, and Saining Xie.
\newblock A {ConvNet} for the 2020s, Mar. 2022.
\newblock arXiv:2201.03545 [cs].

\bibitem{mirsky_creation_and_detection_of_deepfakes_a_survey}
Yisroel Mirsky and Wenke Lee.
\newblock The creation and detection of deepfakes: A survey.
\newblock {\em ACM Comput. Surv.}, 54(1), Jan. 2021.

\bibitem{mubarak_survey_on_the_detection_and_impacts_of_deepfakes}
Rami Mubarak, Tariq Alsboui, Omar Alshaikh, Isa Inuwa-Dutse, Saad Khan, and Simon Parkinson.
\newblock A survey on the detection and impacts of deepfakes in visual, audio, and textual formats.
\newblock {\em IEEE Access}, 11:144497--144529, 2023.

\bibitem{ojha_towards_2023}
Utkarsh Ojha, Yuheng Li, and Yong~Jae Lee.
\newblock Towards {Universal} {Fake} {Image} {Detectors} that {Generalize} {Across} {Generative} {Models}.
\newblock In {\em 2023 {IEEE}/{CVF} {Conference} on {Computer} {Vision} and {Pattern} {Recognition} ({CVPR})}, pages 24480--24489, Vancouver, BC, Canada, June 2023. IEEE.

\bibitem{pirogov2025evaluatingdeepfakedetectorswild}
Viacheslav Pirogov and Maksim Artemev.
\newblock Evaluating deepfake detectors in the wild, 2025.

\bibitem{pu_deepfake_videos_in_the_wild}
Jiameng Pu, Neal Mangaokar, Lauren Kelly, Parantapa Bhattacharya, Kavya Sundaram, Mobin Javed, Bolun Wang, and Bimal Viswanath.
\newblock Deepfake videos in the wild: Analysis and detection.
\newblock In {\em Proceedings of the Web Conference 2021}, WWW '21, page 981–992, New York, NY, USA, 2021. Association for Computing Machinery.

\bibitem{raghu_vision_2022}
Maithra Raghu, Thomas Unterthiner, Simon Kornblith, Chiyuan Zhang, and Alexey Dosovitskiy.
\newblock Do {Vision} {Transformers} {See} {Like} {Convolutional} {Neural} {Networks}?, Mar. 2022.
\newblock arXiv:2108.08810 [cs].

\bibitem{rana_deepfake_2022}
Md~Shohel Rana, Mohammad~Nur Nobi, Beddhu Murali, and Andrew~H. Sung.
\newblock Deepfake {Detection}: {A} {Systematic} {Literature} {Review}.
\newblock {\em IEEE Access}, 10:25494--25513, 2022.

\bibitem{tassone_continuous_fake_media_detection}
Francesco Tassone, Luca Maiano, and Irene Amerini.
\newblock Continuous fake media detection: Adapting deepfake detectors to new generative techniques.
\newblock {\em Computer Vision and Image Understanding}, 249:104143, 2024.

\bibitem{wodajo_deepfake_2021}
Deressa Wodajo and Solomon Atnafu.
\newblock Deepfake {Video} {Detection} {Using} {Convolutional} {Vision} {Transformer}, Mar. 2021.
\newblock arXiv:2102.11126 [cs].

\bibitem{zhang_deepfake_2022}
T. Zhang.
\newblock Deepfake generation and detection, a survey.
\newblock {\em Multimedia Tools and Applications}, 81:6259--6276, February 2022.

\bibitem{zhou_face_forensics_in_the_wild}
Tianfei Zhou, Wenguan Wang, Zhiyuan Liang, and Jianbing Shen.
\newblock Face forensics in the wild.
\newblock {\em CoRR}, abs/2103.16076, 2021.

\bibitem{zi_wilddeepfake}
Bojia Zi, Minghao Chang, Jingjing Chen, Xingjun Ma, and Yu-Gang Jiang.
\newblock Wilddeepfake: A challenging real-world dataset for deepfake detection.
\newblock In {\em Proceedings of the 28th ACM International Conference on Multimedia}, MM '20, page 2382–2390, New York, NY, USA, 2020. Association for Computing Machinery.

\end{thebibliography}
}

\end{document}